%% file: paper_arxiv.tex
\documentclass[sigconf]{acmarxiv}

\usepackage{algorithm}
\usepackage{algorithmic}
\usepackage{pifont}
\usepackage{listings} 
\usepackage{xcolor} 
\usepackage{caption}
\usepackage{mdframed}

\DeclareMathOperator*{\argmin}{argmin}
\DeclareMathOperator*{\argmax}{argmax}

\definecolor{deepgreen}{rgb}{0,0.5,0}

\lstdefinestyle{pythonstyle}{
  language=Python,
  basicstyle=\small \ttfamily,
  keywordstyle=\color{blue},
  commentstyle=\color{gray},
  stringstyle=\color{deepgreen},
  breaklines=true,
  frame=single,
  tabsize=2,
  showstringspaces=false
}
\AtBeginDocument{%
  }

\setcopyright{acmlicensed}
\copyrightyear{2025}
\acmYear{2025}
\acmDOI{XXXXXXX.XXXXXXX}
\acmConference[Preprint]{Under review for International Conference on Knowledge Discovery and Data Mining 2025}{Under Review}{© 2025}

\begin{document}

\title[Cognify: Supercharging Gen-AI Workflows With Hierarchical Autotuning]{Cognify: Supercharging Gen-AI Workflows With Hierarchical Autotuning}

\author{Zijian He}
\authornote{Both authors contributed equally to this research.}
\email{zih015@ucsd.edu}
\author{Reyna Abhyankar}
\authornotemark[1]
\email{vabhyank@ucsd.edu}
\affiliation{%
  \institution{UC San Diego}
  \city{La Jolla}
  \state{CA}
  \country{USA}
}

\author{Vikranth Srivatsa}
\affiliation{%
  \institution{UC San Diego}
  \city{La Jolla}
  \state{CA}
  \country{USA}
}
\email{vsrivatsa@ucsd.edu}

\author{Yiying Zhang}
\affiliation{
  \institution{UC San Diego, GenseeAI, Inc.}
  \city{La Jolla}
  \state{CA}
  \country{USA}
}
\email{yiying@gensee.ai}

\renewcommand{\shortauthors}{He*, Abhyankar*, Srivatsa, Zhang}

\renewcommand{\ttdefault}{lmtt}

\newcommand{\x}{$\times$}
\newcommand{\eg}{e.g.}
\newcommand{\ie}{i.e.}
\newcommand{\sysname}{Cognify}
\newcommand{\search}{\texttt{AdaSeek}}
\newcommand{\fixme}[1]{{\color{red}\textbf{\fbox{FIXME} #1}}}
\newcommand{\FIXME}[1]{{\color{red}\textbf{\fbox{FIXME} #1}}}

\newcommand{\mycap}[2]{\caption{{\bf #1} \textnormal{#2}}}



\sloppy

\input{abstract}




\maketitle

\input{fig-architecture}
\input{introduction}

\input{theory}
\input{design}

\input{results}

\input{related}

\input{conclude}

\bibliographystyle{ACM-Reference-Format}
\bibliography{sysml,all-defs,references}

\input{appendix}

\end{document}

%% file: abstract.tex
\begin{abstract}
Today's gen-AI workflows that involve multiple ML model calls, tool/API calls, data retrieval, or generic code execution are often tuned manually in an ad-hoc way that is both time-consuming and error-prone.
In this paper, we propose a systematic approach for automatically tuning gen-AI workflows.
Our key insight is that gen-AI workflows can benefit from structure, operator, and prompt changes, but unique properties of gen-AI workflows require new optimization techniques.
We propose \search, an adaptive hierarchical search algorithm for autotuning gen-AI workflows. \search\ organizes workflow tuning methods into different layers based on the user-specified total search budget and distributes the budget across different layers based on the complexity of each layer. During its hierarchical search, \search\ redistributes the search budget from less useful to more promising tuning configurations based on workflow-level evaluation results.
We implement \search\ in a workflow autotuning framework called \sysname\ and evaluate \sysname\ using six types of workflows such as RAG-based QA and text-to-SQL transformation.
Overall, \sysname\ improves these workflows' generation 
 quality by up to 2.8\x{}, reduces execution monetary cost by up to 10\x{}, and reduces end-to-end latency by 2.7\x{}. 

\end{abstract}

%% file: fig-architecture.tex
{
\begin{figure*}[t]   
\begin{center}
\centerline{\includegraphics[width=\textwidth]{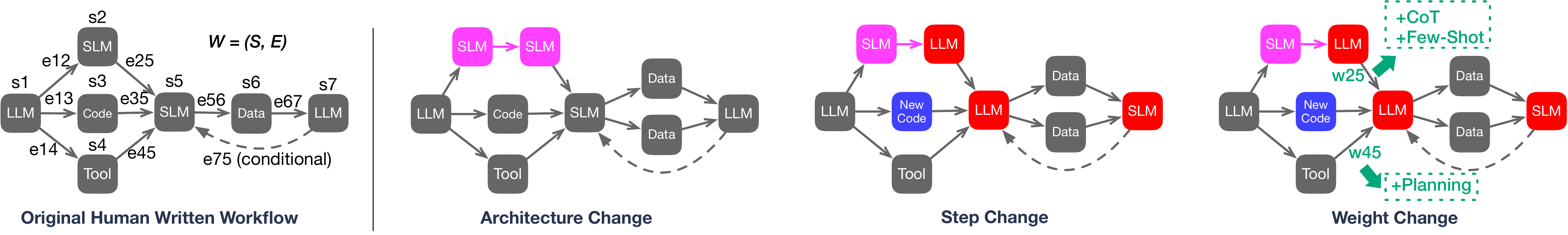}}
\mycap{Gen-AI Workflows Tuning Methods.}
{
SLM and LLM represent different language models (\eg, small and large). Code, Tool, and Data represent code blocks, tool calls, and data retrieval. Dash curved lines represent loops or control flow changes. p* represents prompt optimizations or additional information added for the downstream step.
}
\Description{Four graphs with illustrated optimization methods applied.}
\label{fig-analogy}
\end{center}
\end{figure*}
}

%% file: introduction.tex
\section{Introduction}
\label{sec:intro}

Today's production use of generative AI (gen AI), such as code assistants~\cite{devin}, AI customer service and sales~\cite{salesforce-agentforce}, and AI-assisted search~\cite{vertex_ai_search}, often involves multiple steps of gen-AI model calling, tool/API calling, data retrieval, or generic code execution. 
In this paper, we call such gen-AI software that goes beyond a single model call {\em gen-AI workflows}~\footnote{Terms such as agentic workflows~\cite{andrew_ng_ai_agentic_workflows} and compound AI~\cite{databricks-compound-ai} are also used by others.}.
Gen-AI workflows allow for more customization, integration, and capabilities, thus being the de facto gen-AI product solutions for many business uses.

Despite gen-AI workflows' appeal, their full capabilities are yet to be unlocked due to today's gen-AI workflow practice.
After developing a workflow's basic logic, engineers manually explore various variations to its structure (\eg, by adding a step), test different models, and tune various prompts.
This manual, ad-hoc workflow tuning practice is time-consuming and ineffective, as humans cannot efficiently navigate through a vast search space of tuning options. As a result, gen-AI workflows today still suffer from subpar generation quality and other metrics like end-to-end execution latency. These problems are exacerbated by the rapidly increasing amounts of gen-AI tuning options, such as new models and new prompt engineering techniques~\cite{yao2022react,yao2023tree}.
As more gen-AI workflows are introduced, automated gen-AI workflow optimization is a must. 
Unfortunately, existing automated gen-AI optimizers either focus on single model calls~\cite{routellm,tensor-opera-router} or cater to only a selected set of workflow tuning methods and evaluation metrics~\cite{khattab2024dspy,Trace,textgrad,symbolic}.

We believe workflow auto-tuning should be conducted under a systematic and extensible framework. 
Our insights are that within a workflow, we can view each computation step (a model call, a tool call, a code block, a data retrieval) 
as a node and messages directed from one computation step to another 
as directed edges. The ``weight'' of an edge in a gen-AI workflow is the alteration to the original message sent via the edge (\eg, adding few-shot examples to a prompt message). 
We categorize workflow tuning methods into three types as shown in Figure~\ref{fig-analogy}:
(1) {\em architecture changes} that alter the original workflow by adding/removing/moving steps or edges, such as task decomposition and ensembling,
(2) {\em step changes} that alter the operations performed at a step, such as calling different models or rewriting code pieces,
and (3) {\em weight changes} that alter the value of edges, such as augmentation to prompts.


Although traditional ML models also have similar concepts that manifest as AutoML (or Neural-Architecture Search, NAS) and weight training, classical 
methods like Reinforcement-Learning- 
and Bayesian-Optimization-based NAS~\cite{zoph17nas,falkner2018bohb, kandasamy2018nasbot,ENAS,hutter2019automated} and SGD-based weight training~\cite{lecun1998gradient} do not work for gen-AI workflows for several reasons.
First, unlike ML model weights, ``weights'' in workflows like added prompting messages are not differentiable and thus cannot be trained using SGD. 
Second, workflows (tens of steps and edges) are significantly smaller than ML models, potentially allowing search-based approaches to be applied for all types of workflow tuning changes.
Third, users often have a limited budget (\eg, 50 evaluation iterations) for their workflow optimization, not allowing enough search points for RL to learn from or for traditional BO to converge, especially when considering more types of cogs (\ie, search dimensions).

To confront these challenges, we propose an adaptive, hierarchical search-based algorithm called \textbf{\textit{\search}}. 
\search\ separates architecture, step, and weight changes into different number of search layers based on the user-defined total search budget and adaptively distributes the budget across layers based on each layer's complexity. 
This adaptive budget distribution ensures good search space coverage (with more layers), while allowing for enough local exploration (with fewer layers). 
Within each layer, \search\ uses TPE~\cite{bergstra2011tpe} to sample a small number of configurations at a time, and for each sampled configuration, \search\ recursively samples lower-layer configurations. When all layers have been sampled, \search\ evaluates the workflow with the chosen set of configurations using a user-provided training dataset and evaluator. \search\ stops searching the configurations that yield the lowest $X$\% evaluation results or when the search converges, 
and \search\ redistributes these saved search budget to more promising configurations=.

Based on the \search\ algorithm, we build \textbf{\textit{\sysname}}, a gen-AI workflow optimization framework that systematically and comprehensively optimizes workflows for multiple objectives (final generation quality as defined by a user-provided evaluator, end-to-end request latency, and total workflow execution monetary cost). 
We evaluated \sysname\ with six representative gen-AI workflows, including question-and-answer~\cite{yang2018hotpotqa}, code generation~\cite{humaneval}, data visualization~\cite{datavis}, text-to-SQL generation~\cite{gao2023texttosql}, financial analysis~\cite{finrobot}, and the Big-bench~\cite{bigbench}. We compare \sysname\ with no workflow optimization, DSPy~\cite{DSPy-repo}, and Trace~\cite{Trace}.
Overall, \sysname\ improves workflow generation quality by up to 2.8\x, reduces execution cost by up to 10\x, and reduces end-to-end latency by up to 2.7\x{} compared to original expert-written workflows. \sysname\ outperforms DSPy and Trace with up to 2.6\x{} higher generation quality, up to 10\x{} cost reduction, and up to 3\x{} latency reduction.

Overall, this paper makes the following contributions.

\begin{itemize}
    \item The first formal and comprehensive definition of the gen-AI workflow autotuning problem.
    \item Key insights of how autotuning gen-AI workflows is different from classical AutoML and model weight training.
    \item The \search\ algorithm and its adaptive hierarchical search approach.
    \item The implementation of \sysname, its new programming model (Appendix~\ref{sec:programming-model}), and its full-scale evaluation.
    
\end{itemize}
\sysname\ is open-sourced at \url{https://github.com/GenseeAI/cognify}.

%% file: theory.tex
\section{Formalizing Gen-AI Workflow Autotuning}
\label{sec:theory}

This section presents how we formalize the problem of gen-AI workflow autotuning with a general, extendable framework and how it resonates but differs from classical AutoML \cite{hutter2019automated} and weight training \cite{rumelhart1986learning}.

We formalize the definition of gen-AI workflows as generic program flows that contain at least one generative AI model call. A {\em step}, $s_i$, in a workflow represents a unit whose internal states are not autotuned. For example, a model call is a step as its internal weights are not autotuned at the workflow level. A code block can be a step if it is always tuned as one piece. All steps in a workflow forms the node set, $S$. 
We define an {\em edge}, $e_{ij}$, for every pair of steps $s_i$ and $s_j$ where $s_i$ sends its output to $s_j$. As we allow arbitrary gen-AI workflow program structures, edges in a workflow can represent control paths and form cycles. We define the edge set, $E$. We define the weight of an edge, $w_{ij}$, as the auxilary values or operations applied to the output $e_{ij}$ from $s_i$ to $s_j$. For example, $w_{ij}$ can be few-shot examples added to the output message $o_{ij}$ to form a prompt to the next LLM step, $s_j$. 
A gen-AI workflow, $W$, is $W = (S,E)$.

After a workflow architecture is defined, its structure, step operators, and edge weights can all be altered to achieve better results. We view all these processes as {\em autotuning} a gen-AI workflow and call each type of alteration a {\em cog}. Each cog can have many different values and forms one dimension in the entire autotuning search space. We categorize cogs into three types.
The first type, {\em architecture cogs, $C_a$}, alters the structure of a workflow by adding or removing steps and edges to get a new workflow graph $W_\delta = (S_\delta,E_\delta)$. For example, \sysname\ includes two architecture cogs, task decomposition and task ensembling (Section~\ref{sec:structure-cog}), $c_a^{TD}$ and $c_a^{TE}$. Each cog can be applied to different locations of a workflow, \ie, having different values along their dimensions. 
The second type, {\em step cogs, $C_s$}, alters a step's operation without changing its input and output, \ie, $s_i\rightarrow s_i^{\delta}$. For example, the model cog $c_s^{MD}$ \sysname\ supports uses different models at different model-call steps.
The third method, {\em weight cogs, $C_w$}, alters the values of edges, $e_{ij}$.
For language model steps, their inputs are prompts, and weights are various prompt engineering methods added to the original prompt message.

The effectiveness of gen-AI workflow autotuning is measured by user-defined metrics, $M=(M_1, M_2,..)$. For example, a quality evaluator can be the percentage of workflow generations exactly matching the ground truth. It can also be a numerical score given by an LLM as a judge. Other metrics include end-to-end request execution latency and total workflow execution monetary cost. A workflow is judged by a combination of user-defined metrics with a user-defined evaluator function, $E(M)$. For example, an evaluator can be maximizing the product of quality and cost-effectiveness (the inverse of monetary cost), $E(M_q,M_c)=max(M_q\times \frac{1}{M_c})$. The workflow autotuning process' goal is to apply alterations $(C_a,C_s,C_e)$ to a workflow $W$ to achieve the best $E(M)$.

The above definitions have their similar counterparts in classical ML: ML model architecture search, operator search, weight training, and loss functions.
However, techniques from classical ML do not apply to gen-AI workflow autotuning for several key reasons.
First, architecture, step, and weight cogs are non-differentiable, and metrics to evaluate workflows are often discrete and multi-dimensional.
Thus, unlike classical ML that performs weight training using SGD, all types of cogs in workflow autotuning including weight cogs could use discrete search methods.
Second, gen-AI workflows usually have no more than tens of steps and edges, orders of magnitude smaller than modern ML models that have billions of parameters.
The search space of cogs is significantly smaller than the search space in traditional AutoML, allowing for search-based approaches like Bayesian Optimization.
Third, alterations to a workflow (\ie, cog values) are sometimes non-structured and have no known super set, as some cog values can be proposed by an LLM. Thus, traditional NAS approaches based on super-models~\cite{ENAS} do not work for workflows.
Finally, unlike traditional AutoML and weight training, gen-AI workflows often need to be tuned more frequently because of the fast development of gen-AI models and their use cases. As each workflow autotuning is expected to incur a monetary burden, users are likely to use relatively small search budgets (\eg, tens to hundreds of iterations). Small budgets render RL-based search solutions infeasible, as they cannot acquire enough experience. 


%% file: design.tex
\input{search-algo}

\input{cognify}

%% file: search-algo.tex
\section{The \search\ Search Algorithm}
\label{sec:search}



With our insights in Section~\ref{sec:theory}, we believe that search methods based on Bayesian Optimizer (BO) can work for all types of cogs in gen-AI workflow autotuning because of BO's efficiency in searching discrete search space.
A key challenge in designing a BO-based search is the limited search budgets that need to be used to search a high-dimensional cog space. 
For example, for 4 cogs each with 4 options and a workflow of 3 LLM steps, the search space is $4^{12}$. Suppose each search uses GPT-4o and has 1000 output tokens, the entire space needs around \$168K to go through. A user search budget of \$100 can cover only 0.06\% of the search space. A traditional BO approach cannot find good results with such small budgets.

To confront this challenge, we propose \textit{\textbf{\search}}, an adaptive hierarchical search algorithm that efficiently assigns search budget across cogs based on budget size and observed workflow evaluation results, as defined in Algorithms~\ref{alg:main} and \ref{alg:outer} and described below.



\input{algo-main-search}

\subsection{Hierarchical Layer and Budget Partition}
\label{sec:ssp}

A non-hierarchical search has all cog options in a single-layer search space for an optimizer like BO to search, an approach taken by prior workflow optimizers~\cite{dspy-2-2024,gptswarm}.
With small budgets, a single-layer hierarchy allows BO-like search to spend the budget on dimensions that could potentially generate some improvements.
However, a major issue with a single-layer search space is that a search algorithm like BO can be stuck at a local optimum even when budgets increase.
To mitigate this issue, our idea is to perform a hierarchical search that works by choosing configurations in the outermost layer first, then under each chosen configuration, choosing the next layer's configurations until the innermost layer. 
With such a hierarchy, a search algorithm could force each layer to sample some values. Given enough budget, each dimension will receive some sampling points, allowing better coverage in the entire search space. However, with high dimensionality (\ie, many types of cogs) and insufficient budget, a hierarchical search may not be able to perform enough local search to find any good optimizations.

To support different user-specified budgets and to get the best of both approaches, we propose an adaptive hierarchical search approach, as shown in Algorithm~\ref{alg:main}.
\search\ starts the search by combining all cogs into one layer ($L=1$, line 9 in Algorithm~\ref{alg:main}) and estimating the expected search budget of this single layer to be the total number of cogs to the power of $\alpha$ (lines 16-19, by default $\alpha = 1.1$). This budget is then passed to the \texttt{LayerSearch} function (Algorithm~\ref{alg:outer}) to perform the actual cog search. When the user-defined budget is no larger than this estimated budget, we expect the single-layer, non-hierarchical search to work better than hierarchical search.

If the user-defined budget is larger, \search\ continues the search with two layers ($L=2$), combining step and weight cogs into the inner layer and architecture cogs as the outer layer (lines 11-14).
\search\ estimates the total search budget for this round as the product of the number of cogs in each of the two layers to the power of $\alpha$ (lines 16-20). It then distributes the estimated search budget between the two layers proportionally to each layer's complexity (lines 22-24) and calls the upper layer's \texttt{LayerSearch} function. Afterward, if there is still budget left, \search\ performs a last round of search using three layers and the remaining budget in a similar way as described above but with three separate layers (architecture as the outermost, step as the middle, and weight cogs as the innermost layer). Two or three layers work better for larger user-defined budgets, as they allow for a larger coverage of the high-dimensional search space.

Finally, \search\ combines all the search results to select the best configurations based on user-defined metrics (line 34).

\subsection{Recursive Layer-Wise Search Algorithm}
We now introduce how \search\ performs the actual search in a recursive manner until the inner-most layer is searched, as presented in Algorithm~\ref{alg:outer} \texttt{LayerSearch}. 
Our overall goal is to ensure strong cog option coverage within each layer while quickly directing budgets to more promising cog options based on evaluation results.
Specifically, every layer's search is under a chosen set of cog configurations from its upper layers ($C_{chosen}$) and is given a budget $b$. 
In the inner-most layer (lines 7-20), \search\ samples $b$ configurations and evaluates the workflow for each of them together with the configurations from all upper layers ($C_{chosen}$). The evaluation results are added to the feedback set $F$ as the return of this layer.

\input{algo-outer-search}

For a non-inner-most layer, \search\ samples a chunk ($W$) of points at a time using the TPE BO algorithm~\cite{bergstra2011tpe} until all this layer's pre-assigned budget is exhausted (lines 27-30). Within a chunk, \search\ uses a successive-halving-like approach to iteratively direct the search budget to more promising configurations within the chunk (the dynamically changing set, $\Theta$). In each iteration, \search\ calls the next-level search function for each sampled configuration in $\Theta$ with a budget of $r_s$ and adds the evaluation observations from lower layers to the feedback set $F$ for later TPE sampling to use (lines 35-37).
In the first iteration ($s=0$), $r_s$ is set to $R\cdot \eta^0=R$ (line 34). After the inner layers use this budget to search, \search\ filters out configurations with lower performance and only keeps the top $\lfloor \frac{|\Theta|}{\eta}\rfloor$ configurations as the new $\Theta$ to explore in the next iteration (line 42). In each next iteration, \search\ increases $r_s$ by $\eta$ times (line 34), essentially giving more search budget to the better configurations from the previous iteration.

The successive halving method effectively distributes the search budget to more promising configurations, while the chunk-based sampling approach allows for evaluation feedback to accumulate quickly so that later rounds of TPE can get more feedback (compared to no chunking and sampling all $b$ configurations at the same time). To further improve the search efficiency, we adopt an {\em early stop} approach where we stop a chunk or a layer's search when we find its latest few searches do not improve workflow results by more than a threshold, indicating convergence (lines 14,38,45).

%% file: algo-main-search.tex
\begin{algorithm}[h]
    \caption{\search\ Algorithm}
    \label{alg:main}
      \small
\begin{algorithmic}[1]
\STATE \textbf{Global Value:} $R = \emptyset$ \COMMENT{Global result set}

\STATE \textbf{Input:} User-specified Total Budget $TB$
\STATE \textbf{Input:} Cog set $C = \{c_{11},c_{12},...\}, \{c_{21},c_{22},...\}, \{c_{31},c_{32},...\}$

    \STATE
\STATE $U = 0$ \COMMENT{Used budget so far, initialize to 0}

\STATE \COMMENT{Perform search with 1 to 3 layers until budget runs out}
\FOR{$L = 1,2,3$} 
        \IF{$L=1$}
            \STATE $C_1 = C_1 \cup C_2 \cup C_3$ \COMMENT{Merge all cogs into a single layer}
        \ENDIF
        \IF{$L==2$}
            \STATE $C_1 = C_1 \cup C_2$ \COMMENT{Merge step and weight cogs}
            \STATE $C_2 = C_3$ \COMMENT{Architecture cog becomes the second layer}
        \ENDIF
        \STATE
    \FOR{$i = 1,..,L$}
    \STATE $NC_i = |C_i|$ \COMMENT{Total number of cogs in layer $L$} 
    \STATE $S_i = NC_i^\alpha$ \COMMENT{Estimated expected search size in layer $i$}
    \ENDFOR
    \STATE $E_L = \prod\limits_{i=1}^{L}S_i$ \COMMENT{Expected total search size in the current round}
    \STATE $E = TB - U > E_L$ ? $E_L$ : $(TB - U)$ \COMMENT{Consider insufficient budget} 
    \IF{$L==3$ and $(TB - U)$ > $E_L$}
         \STATE $E = TB - U$ \COMMENT{Spend all remaining budget if at 3 layer}
    \ENDIF
    \FOR{$i = 1,..,L$}
        \STATE $B_i =  \lfloor S_i \times \sqrt[L]{\frac{E}{E_L}}\rfloor$
        \COMMENT{Assign budget proportionally to $S_i$}
    \ENDFOR
    \STATE
\STATE \texttt{LayerSearch} ($\emptyset$, $B$, $L$, $B_L$) \COMMENT{Hierarchical search from layer $L$}
\STATE
\STATE $U = U + E$
\IF{$U \geq TB$}
\STATE break \COMMENT{Stop search when using up all user budget}
\ENDIF
\ENDFOR
\STATE
\STATE \textbf{Output:} $O$ = \texttt{SelectBestConfigs} ($R$) \COMMENT{Return best optimizations}
\end{algorithmic}
\end{algorithm}

%% file: algo-outer-search.tex
\begin{algorithm}[h]
  \small
    \caption{\texttt{LayerSearch} Function}
    \label{alg:outer}
\begin{algorithmic}[1]
\STATE \textbf{Global Value:} $R$ \COMMENT{Global result set}
\STATE \textbf{Input:} $C_{chosen}$: configs chosen in upper layers
\STATE \textbf{Input:} $B$: Array storing assigned budgets to different layers
\STATE \textbf{Input:} $curr\_layer$: this layer's level
\STATE \textbf{Input:} $curr\_b$: this layer's assigned budget

    \STATE
    \STATE \COMMENT{Search for inner-most layer}
    \IF{curr\_layer == 1}
        \STATE $F = \emptyset$ \COMMENT{Init this layer's feedback set to empty}
        \FOR{$k = 0, \dots, curr\_b$}
            \STATE $\lambda$ = \texttt{TPESample} (1) \COMMENT{Sample one configuration using TPE}
            \STATE $f = $ \texttt{EvaluateWorkflow} ($C_{chosen} \cup \lambda$)
            \STATE $R = R \cup \{C_{chosen} \cup \lambda\}$ \COMMENT{Add configuration to global $R$}
            \IF{\texttt{EarlyStop} (f)}
            \STATE break
            \ENDIF
            \STATE $F = F \cup \{f\}$ \COMMENT{Add evaluate result to feedback $F$}
        \ENDFOR
        \STATE \textbf{Return} $F$
    \ENDIF
    \STATE
    \STATE \COMMENT{Search for non-inner-most layer}
    \STATE $b\_used = 0$, $TF = \emptyset$ \COMMENT{Init this layer's used budget and feedback set}
    \STATE $R = \lceil\frac{curr\_b}{\eta}\rceil$, $S = \lfloor\frac{curr\_b}{R}\rfloor$ \COMMENT{Set $R$ and $S$ based on $curr\_b$}
    \STATE
    \WHILE{$b\_{used}$ $\leq$ $curr\_b$}
        \STATE \COMMENT{Sample $W$ configs at a time until running out of $curr\_b$}
        \STATE $n = (curr\_b - b_{used})$ > $W$ ? $W$ : $(curr\_b - b_{used})$
        \STATE $b\_used$ += $n$
        \STATE $\Theta = $ \texttt{TPESample} ($n$) \COMMENT{Sample a chunk of $n$ configs in the layer} 
        \STATE $F = \emptyset$ \COMMENT{Init this layer's feedback set to empty}
        \STATE
        \FOR{$s = 0, 1, \dots, S$}
            \STATE $r_s = R\cdot \eta^s$
            \FOR{$\theta \in \Theta$}
                    \STATE $f =$ \texttt{LayerSearch} ($C_{chosen} \cup \{\theta\}$, $B$, curr\_layer$-1$, $r_s$)
                \STATE $F = F \cup f$ \COMMENT{Add evaluate result to feedback}
                \IF{\texttt{EarlyStop} ($f$)}
                    \STATE $\Theta = \Theta - \{\theta\}$ \COMMENT{Skip converged configs}
                \ENDIF
            \ENDFOR
            \STATE $\Theta$ = Select top $\lfloor \frac{|\Theta|}{\eta}\rfloor$ configs from $F$ based on user-specified metrics
        \ENDFOR
        \STATE
        \IF{\texttt{EarlyStop} ($F$)}
            \STATE break \COMMENT{Skip remaining search if results converged}
        \ENDIF
        \STATE $TF = TF \cup F$
    \ENDWHILE
        \STATE \textbf{Return} $TF$


\end{algorithmic}
\end{algorithm}






%% file: cognify.tex
\section{\sysname\ Design}
\label{sec:cognify}

We build \sysname, an extensible gen-AI workflow autotuning platform based on the \search\ algorithm. The input to \sysname\ is the user-written gen-AI workflow (we currently support LangChain \cite{langchain-repo}, DSPy \cite{khattab2024dspy}, and our own programming model), a user-provided workflow training set, a user-chosen evaluator, and a user-specified total search budget. \sysname\ currently supports three autotuning objectives: generation quality (defined by the user evaluator), total workflow execution cost, and total workflow execution latency. Users can choose one or more of these objectives and set thresholds for them or the remaining metrics (\eg, optimize cost and latency while ensuring quality to be at least 5\% better than the original workflow). 
\sysname\ uses the \search\ algorithm to search through the cog space.
When given multiple optimization objectives, \sysname\ maintains a sorted optimization queue for each objective and performs its pruning and final result selection from all the sorted queues (possibly with different weighted numbers).
To speed up the search process, we employ parallel execution, where a user-configurable number of optimizers, each taking a chunk of search load, work together in parallel. 
\sysname\ returns multiple autotuned workflow versions based on user-specified objectives.
\sysname\ also allows users to continue the auto-tuning from a previous optimization result with more budgets so that users can gradually increase their search budget without prior knowledge of what budget is sufficient.
Appendix~\ref{sec:apdx-example} shows an example of \sysname-tuned workflow outputs. 
\sysname\ currently supports six cogs in three categories, as discussed below. 


\subsection{Architecture Cogs}
\label{sec:structure-cog}
\sysname\ currently supports two architecture cogs: task decomposition and task ensemble.
Task decomposition~\cite{khot2023decomposed} breaks a workflow step into multiple sub-steps and can potentially improve generation quality and lower execution costs, as decomposed tasks are easier to solve even with a small (cheaper) model.
There are numerous ways to perform task decomposition in a workflow. 
To reduce the search space, we propose several ways to narrow down task decomposition options. Even though we present these techniques in the setting of task decomposition, they generalize to many other structure-changing tuning techniques.

Intuitively, complex tasks are the hardest to solve and worth decomposition the most. We use a combination of LLM-as-a-judge \cite{vicuna_share_gpt} and static graph (program) analysis to identify complex steps. We instruct an LLM to give a rating of the complexity of each step in a workflow. We then analyze the relationship between steps in a workflow and find the number of out-edges of each step (\ie, the number of subsequent steps getting this step's output). More out-edges imply that a step is likely performing more tasks at the same time and is thus more complex. We multiply the LLM-produced rating and the number of out-edges for each step and pick the modules with scores above a learnable threshold as the target for task decomposition. We then instruct an LLM to propose a decomposition (\ie, generate the submodules and their prompts) for each candidate step. 

\input{fig-all-in-one}

The second structure-changing cog that \sysname\ supports is task ensembling. This cog spawns multiple parallel steps (or samplers) for a single step in the original workflow, as well as an aggregator step that returns the best output (or combination of outputs). By introducing parallel steps, \sysname\ can optimize these independently with step and weight cogs. This provides the aggregator with a diverse set of outputs to choose from. 

\subsection{Step Cogs}
We currently support two step-changing cogs: model selection for language-model (LM) steps and code rewriting for code steps.

For model selection, to reduce its search space, we identify ``important'' LM steps---steps that most critically impact the final workflow output to reduce the set \search\ performs TPE sampling on. Our approach is to test each step in isolation by freezing other steps with the cheapest model and trying different models on the step under testing. 
We then calculate the difference between the model yielding the best and worst workflow results as the importance of the step under testing. 
After testing all the steps, we choose the steps with the highest K\% importance as the ones for TPE to sample from.

The second step cog \sysname\ supports is code rewriting, where it automatically changes code steps to use better implementation. To rewrite a code step, \sysname\ finds the $k$ worst- and best-performing training data points and feeds their corresponding input and output pairs of this code step to an LLM. We let the LLM propose $n$ new candidate code pieces for the step at a time.
In subsequent trials, the optimizer dynamically updates the candidate set using feedback from the evaluator.

\subsection{Weight Cogs}
\sysname\ currently supports two weight-changing cogs: reasoning and few-shot examples.
First, \sysname\ supports adding reasoning capability to the user's original prompt, with two options: zero-shot Chain-of-Thought \cite{wei2022chain} (\ie, ``think step-by-step...'') and dynamic planning \cite{huang2022language} (\ie, ``break down the task into simpler sub-tasks...''). These prompts are appended to the user's prompt. In the case where the original module relies on structured output, we support a reason-then-format option that injects reasoning text into the prompt while maintaining the original output schema.

Second, \sysname\ supports dynamically adding few-shot examples to a prompt. At the end of each iteration, we choose the top-$k$-performing examples for an LM step in the training data and use their corresponding input-output pairs of the LM step as the few-shot examples to be appended to the original prompt to the LM step for later iterations' TPE sampling. As such, the set of few-shot examples is constantly evolving during the optimization process based on the workflow's evaluation results. 

%% file: fig-all-in-one.tex
{
\begin{figure*}[t!]
\begin{center}
\centerline{\includegraphics[width=0.95\textwidth]{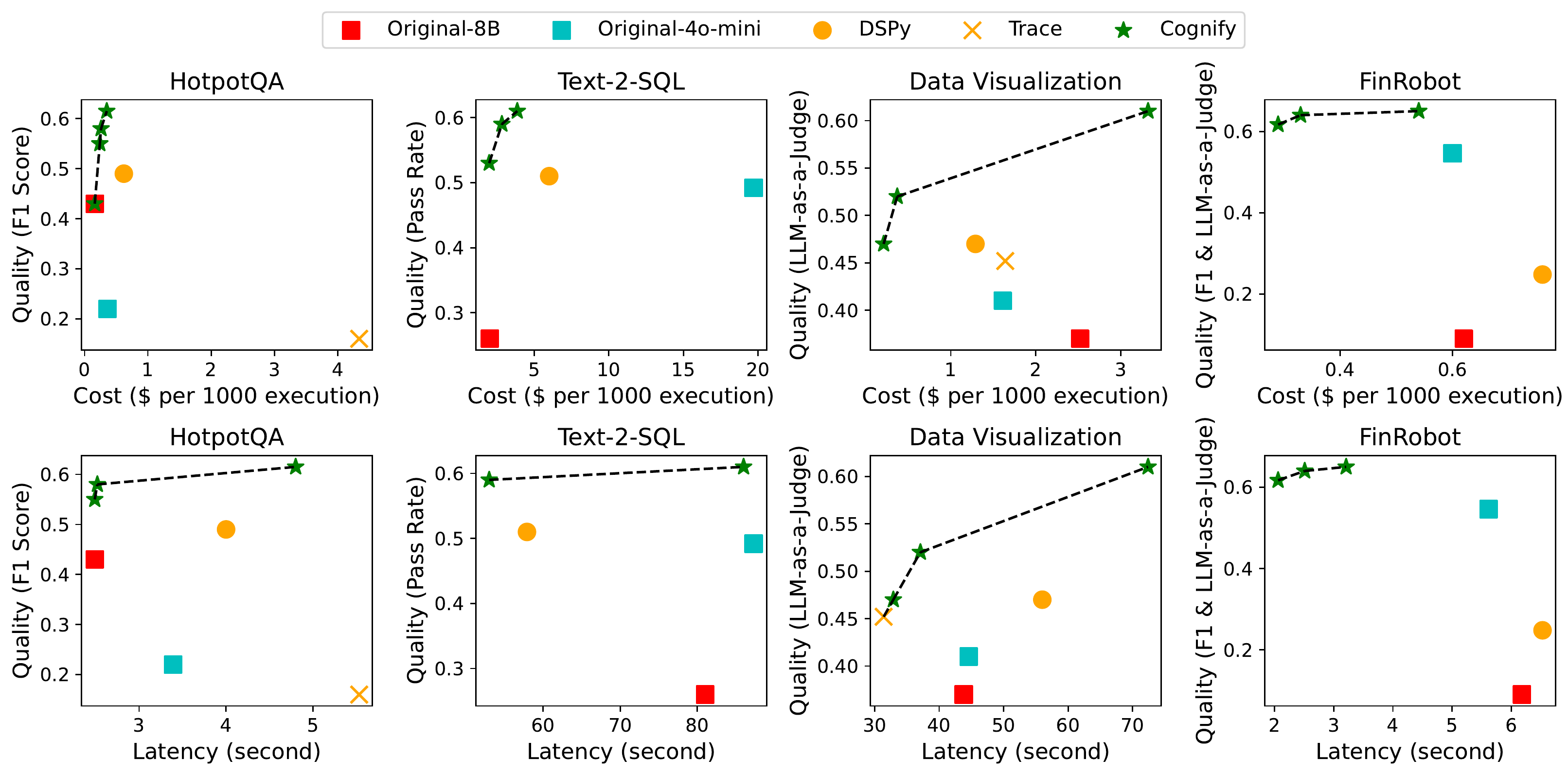}}
\vspace{-0.1in}
\mycap{Generation Quality vs Cost/Latency.}{Dashed lines show the Pareto frontier (upper left is better). Cost shown as model API dollar cost for every 1000 requests. Cognify selects models
from GPT-4o-mini and Llama-8B. DSPy and Trace do not support model selection and are given GPT-4o-mini for all steps. Trace results for Text-2-SQL and FinRobot have 0 quality and are not included.} 
\Description{Eight graphs with different shapes representing baselines compared to points on a Pareto frontier.}
\label{fig-biggrid}
\end{center}
\end{figure*}
}

%% file: results.tex
\section{Evaluation Results}
\label{sec:results}

We evaluate Cognify across six types of tasks: question-answering (HotPotQA~\cite{yang2018hotpotqa}), text-to-SQL (BIRD~\cite{gao2023texttosql}), data visualization (MatPlotAgent benchmark~\cite{datavis}), financial analysis (FinRobot \cite{finrobot}), code generation (HumanEval~\cite{humaneval}), and BigBench~\cite{bigbench}. The quality evaluators used in these tasks, respectively, are F1 score (text similarity to the ground truth answer), pass rate (exactly correct SQL results out of all test data), LLM-as-a-Judge, a combination of F1-score and LLM-as-a-judge, pass rate (exactly correct code results out of all test data), and exact match rate (for word sorting and object counting). Out of these workflows: FinRobot employs a leader-agent that routes requests to other different workers; text-to-SQL and data visualization both loop over model calls for iterative refinement; question-answering involves multi-step retrieval-augmented generation (RAG) \cite{lewis20rag}. 

\input{tbl-bbh}

We compare \sysname\ with three baselines: the original workflows from open-source repositories, DSPy~\cite{DSPy-repo}, and Trace~\cite{Trace}. 
Unless otherwise specified, we give \sysname, DSPy, and Trace all a search budget of 64. \sysname\ chooses from two language models for its model-step cog: GPT-4o-mini~\cite{4o-mini} (via OpenAI API endpoints~\cite{openaiapi}) and Llama-8B~\cite{llama3.1} (via Fireworks API endpoints~\cite{fireworksai2024api}). As the baselines do not have automated model selection capabilities, we run each of them fully on GPT-4o-mini or all on Llama-8B.

\subsection{Overall Workflow Optimization Results}

We first present the overall results across all workloads, highlighting the first four in Figure~\ref{fig-biggrid}.
Overall, \sysname\ improves generation quality (by up to 2.8\x{}), reduces execution cost (by up to 10\x{}), and execution latency (by up to 2.7\x{}) compared to the original workflows across all the workloads by pushing their Pareto frontiers. 
\sysname\ also improves quality, cost, and latency over DSPy and Trace across workloads.

Comparing across workloads, \sysname\ has the largest improvements on HotpotQA (2.8\x{} over the original workflow using 4o-mini) and data visualization (38\% higher than the original workflow using only 4o-mini or only Llama 3.1-8B). \sysname\ achieves its quality improvement for HotpotQA due to its good selection of few-shot examples at various edges. \sysname\ achieves its benefit for data visualization from inserting chain-of-thought reasoning at the beginning of the workflow and planning steps during the initial code generation phase. 
\sysname\ has the largest cost saving on text-to-SQL (10\x{} cheaper) by introducing reasoning; the original workflow included a significant number of retries that \sysname\ avoids with improved generation in earlier steps of the workflow. \sysname\ has the largest latency cut for FinRobot (2.5\x{} faster) because the model selection cog chooses the faster Llama-8B model for certain steps.

For all workloads, DSPy performs better than Trace, although still worse than \sysname. This is because DSPy generates few-shot examples during its optimization process, whereas Trace primarily relies on rewriting user-annotate code blocks. While code rewriting may be effective to generate Trace is especially ineffective for the Text-2-SQL and FinRobot workloads, as its optimized workflow yields 0 quality, \ie, no generated SQL queries / analyses are correct. This is because Trace has a strong tendency to overfit to specific training examples or generate erroneous code rewrites.
DSPy is also not as effective on the FinRobot workload, as their optimized workflows have worse results than the original workflow running GPT-4o-mini. DSPy is unable to combine complex prompt optizations (\eg reasoning) with structured output generation, which is required by the FinRobot task.  

\input{fig-budget-layer-metrics}
\input{fig-search-trace}

We use three additional benchmarks, code generation from HumanEval, word sorting from BigBench, and object counting from BigBench, to demonstrate Cognify's ability to focus on a single objective, as DSPy and Trace are both designed for quality improvement. The evaluator in these tasks is pass rate (either the workflow output is correct or incorrect) or exact match. This forces a stricter quality expectation. \sysname\ improves the accuracy of code gen over the original workflow by 30\% and over Trace and DSPy by 4\%. 
On BBH-object counting, \sysname\ demonstrates an impressive 95\% accuracy, which is 9\% higher than Trace and 2.2\x{}\ higher than DSPy. On BBH-word sorting, \sysname\ is near-perfect, achieving 99\% accuracy, which is 11\% higher than Trace. Both DSPy (on word sorting) and the original workflow (on word sorting and objecting counting) are unable to generate the answer that matches exactly with the expected output. \sysname's code rewriter is primarily responsible for the improvement in quality on these workflows.

\subsection{Detailed Evaluation Results}

We now explain \sysname's benefits and sensitivity with more detailed experiments.

\subsubsection{Search Effectiveness}

\input{fig-input-sensitivity}
\paragraph{Comparison to grid search.}
To evaluate the effectiveness of \sysname's search, we perform an exhaustive grid search of 4096 configurations for the HotpotQA question-answering workflow.
Figure~\ref{fig-searchtrace} plots the grid search results and \sysname's 128 iteration results. As indicated by the heatmap, \sysname\ quickly moves towards the highest quality and lowest cost in approximately 20 iterations. With just 1/32 of the grid search budget, \sysname\ finds 5 new points on the Pareto frontier and misses only one point found by grid search. 

\paragraph{Layering and search budgets.}

We validate our hypothesis that an increasing budget should correspond to an increase in the number of layers used by \sysname. Figure~\ref{fig-layering-influence} shows the effect of using different number of layers with different budgets across all metrics on the text-to-SQL workload. When the budget is small (\ie, 16 iterations), a single layer will yield the best results due to faster convergence to a local optimum. As budget increases, layering allows a more diverse exploration of the search space. At 64 iterations, a 2-layer approach performs the best, and at 128 iterations (our maximum budget used for experiments), 3 layers is superior. This pattern holds across quality, cost, and latency independently.

\paragraph{Sensitivity to training input size.}
We evaluate the effectiveness of \search\ with varying training input sizes on the FinRobot workload in Figure~\ref{fig-input-sensitivity}. With just 6 or more examples, \search\ is able to find higher quality optimized workflows. In less than 50 examples, \search\ generates a 13\% improvement in quality and a 19\% improvement in less than 100 examples. This mirrors the size of typical gold datasets in production workflows \cite{kulkarni25gold}. The only instance in which the search degrades in quality relative to the original workflow is when the training dataset contains only 3 examples, which is an unlikely scenario for workflow developers.

\input{tbl-comparison}

\input{fig-ablation}
\subsubsection{Ablation Study}
To understand where \sysname's benefits come from, we evaluate the effect of different techniques by adding one at a time on the Text-to-SQL workload, as shown in Figure~\ref{fig-ablation-study}. 
The initial version is a non-hierarchical search approach that places all cogs in a single layer.
We then incorporate our adaptive hierarchical layer approach and budget distribution, which results in improved quality configurations after 24 search iterations.
Afterward, we add the per-layer chunk-based successive-halving approach, yielding additional quality improvements as seen by the green line.
Finally, we enable early exiting, achieving the best quality as shown in the red line.

\subsubsection{Search Time and Cost}

We compare the optimization cost and time of \sysname\ and DSPy in Figure~\ref{fig-opt-time-cost}. 
\sysname\ completes its optimization in 1.7\x{} to 2.5\x{} less time than DSPy due to \sysname's efficient use of the overall training budget and its parallel search mechanism. 
While DSPy's optimizer can be marginally cheaper in \$ cost for smaller workflows, its cost bloats for more complex workflows. For example, the cost of optimizing text-to-SQL with DSPy is \$24, which is 2.4\x{} more expensive than \sysname.




%% file: tbl-bbh.tex
\begin{table}[h!]
\centering

\begin{tabular}{|c|c|c|c|c|}
\hline
\textbf{}            & \textbf{Original} & \textbf{DSPy} & \textbf{Trace} & \textbf{Cognify} \\ \hline
\textbf{Code Gen}     & 70\%              & 87\%          & 87\%           & \textbf{91\%}             \\ \hline
\textbf{Word Sorting} & 0\%               & 42\%           & 89\%           & \textbf{99\%}             \\ \hline
\textbf{Object Counting}     & 0\%               & 0\%          & 87\%           & \textbf{95\%}             \\ \hline
\end{tabular}
\label{tab:comparison}
\vspace{0.1in}
\mycap{Accuracy of Code Generation, Word Sorting, and Object Counting.}{}
\end{table}

%% file: fig-budget-layer-metrics.tex
{
\begin{figure*}[t!]
\begin{center}
\centerline{\includegraphics[width=0.95\textwidth]{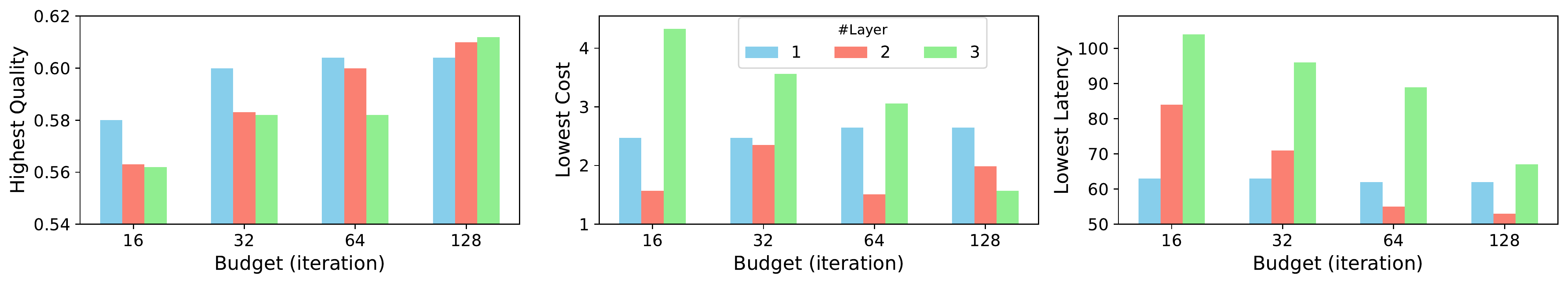}}
\vspace{-0.1in}
\mycap{Effectiveness of Layering over Budget.}{The quality (higher is better), cost (lower is better), and latency (lower is better) achieved on Text-to-SQL by \sysname\ when using different number of search layers under different budget.}
\Description{Three bar graphs (one for each metric), each with three bars that represent layering strategy.}
\label{fig-layering-influence}
\end{center}
\end{figure*}
}

%% file: fig-search-trace.tex
{
\begin{figure}[htbp]
\begin{center}
\centerline{\includegraphics[width=\columnwidth]{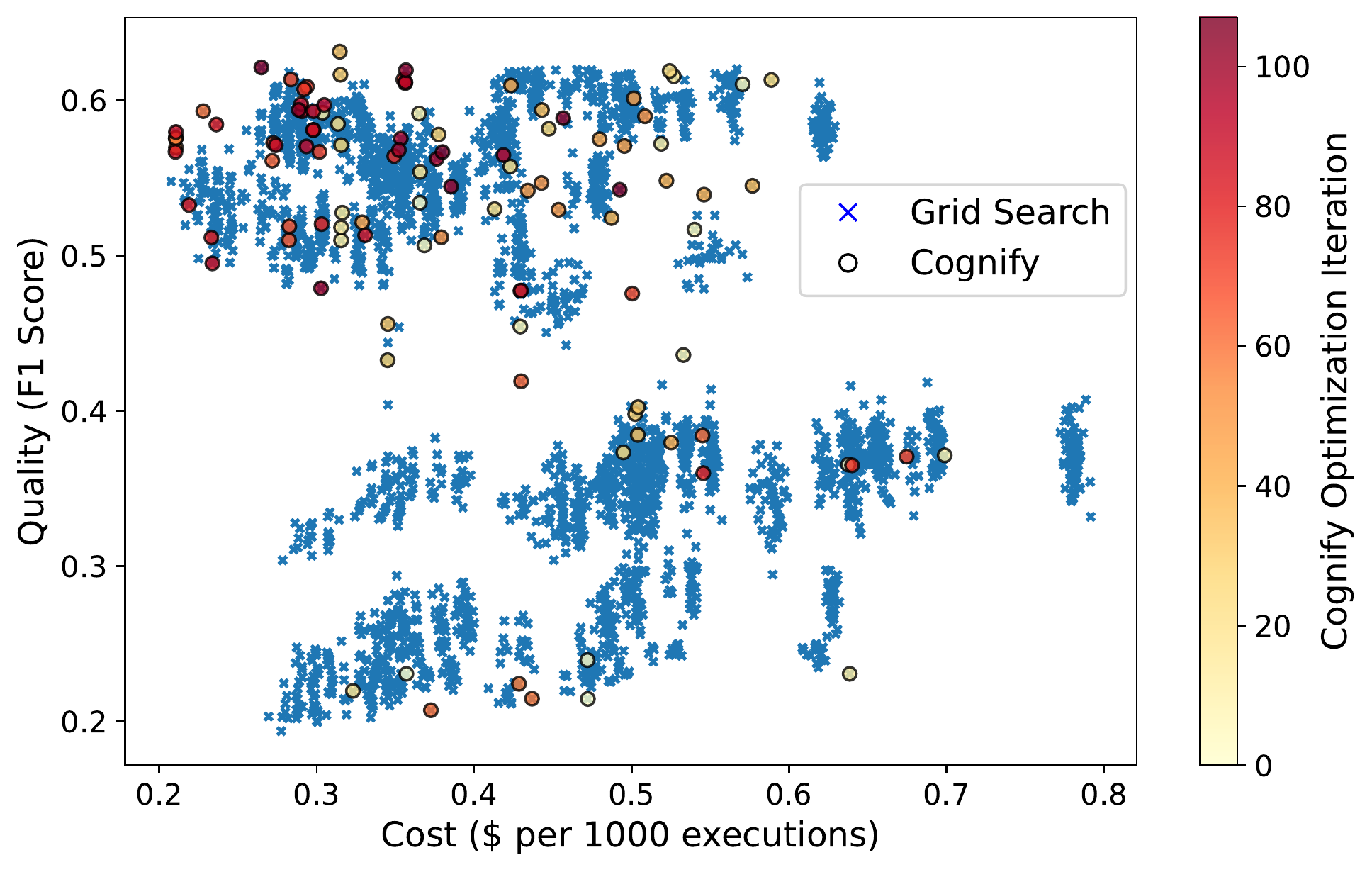}}
\vspace{-0.1in}
\mycap{Grid Search and Cognify's Searched Configurations.}
{
Performed on the HotpotQA workload. Cognify's search results are ordered by iterations from yellow to red colors (up to 128 iterations). Grid search explores the entire 4096 configurations. 
}
\Description{A scatter plot representing a grid search and heatmap of points mapped on top of the grid.}
\label{fig-searchtrace}
\end{center}
\end{figure}
}


%% file: fig-input-sensitivity.tex
{
\begin{figure}[h]
\centerline{\includegraphics[width=0.85\columnwidth]{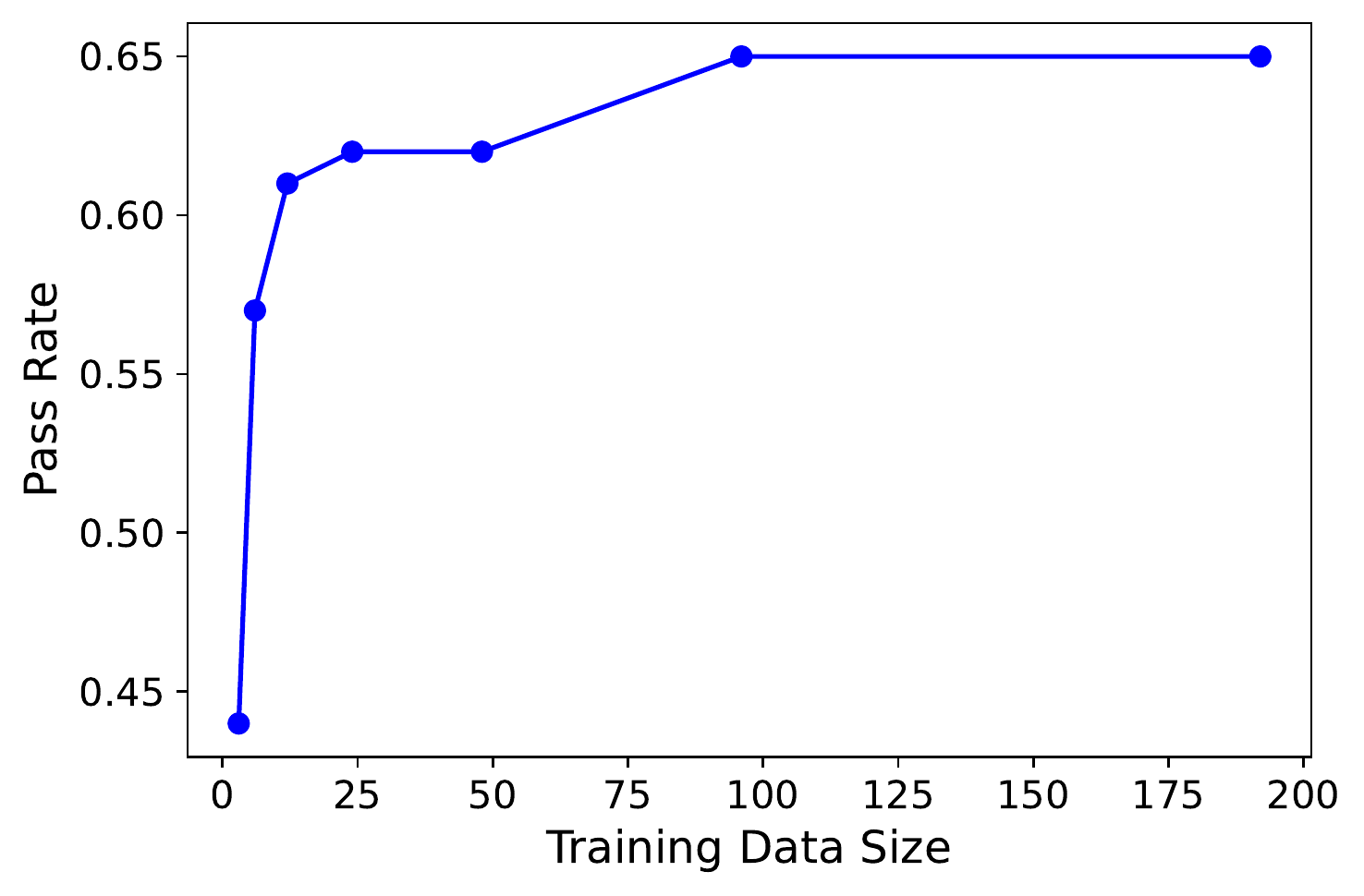}}
\vspace{-0.1in}
\mycap{Sensitivity to Training Set Size.} {Conducted on FinRobot. X-axis represents number of training data points, Y-axis represents quality.}
\Description{A line graph that increases initially before plateauing.}
\label{fig-input-sensitivity}
\end{figure}
}

%% file: tbl-comparison.tex
\newcommand{\cmark}{\ding{51}}
\newcommand{\cmarkbold}{\ding{52}}
\newcommand{\xmark}{\ding{55}}

\begin{table*}[h]
    \centering
    \small
    \begin{tabular}{|l|c|c|c|c|c|l|}
        \hline
        \textbf{System} & \textbf{Structure} & \textbf{Step Change} & \textbf{Prompt Tuning} & \textbf{Multi-objective} & \textbf{Extensible} & \textbf{Optimize Technique} \\
        \hline
        TextGrad, OPRO  & \xmark & \xmark & \cmark & \xmark & \xmark & LLM guided \\
        Symbolic  & \cmark & \xmark & \cmark & \xmark & \xmark & LLM guided \\
        Trace     & \xmark & \cmark\ (Code) & \cmark & \xmark & \xmark & LLM guided \\
        DSPy      & \xmark & \xmark & \cmark & \xmark & \xmark* & Non-hierarchical search \\
        GPTSwarm      & \cmark & \xmark & \cmark & \xmark & \xmark & Non-hierarchical search \\
        \textbf{Cognify}   & \cmarkbold & \cmarkbold\ (Model and Code) & \cmarkbold & \cmarkbold & \cmarkbold & \textbf{Hierarchical search} \\
        \hline
    \end{tabular} 
\vspace{0.1in}
    \mycap{Comparison of Different Gen-AI Workflow Optimizers.} {*DSPy adds new optimizer for each tuning mechanism.}
    \label{tbl-comparison}
\end{table*}

%% file: fig-ablation.tex
{
\begin{figure}[h]
\begin{center}
\centerline{\includegraphics[width=0.9\columnwidth]{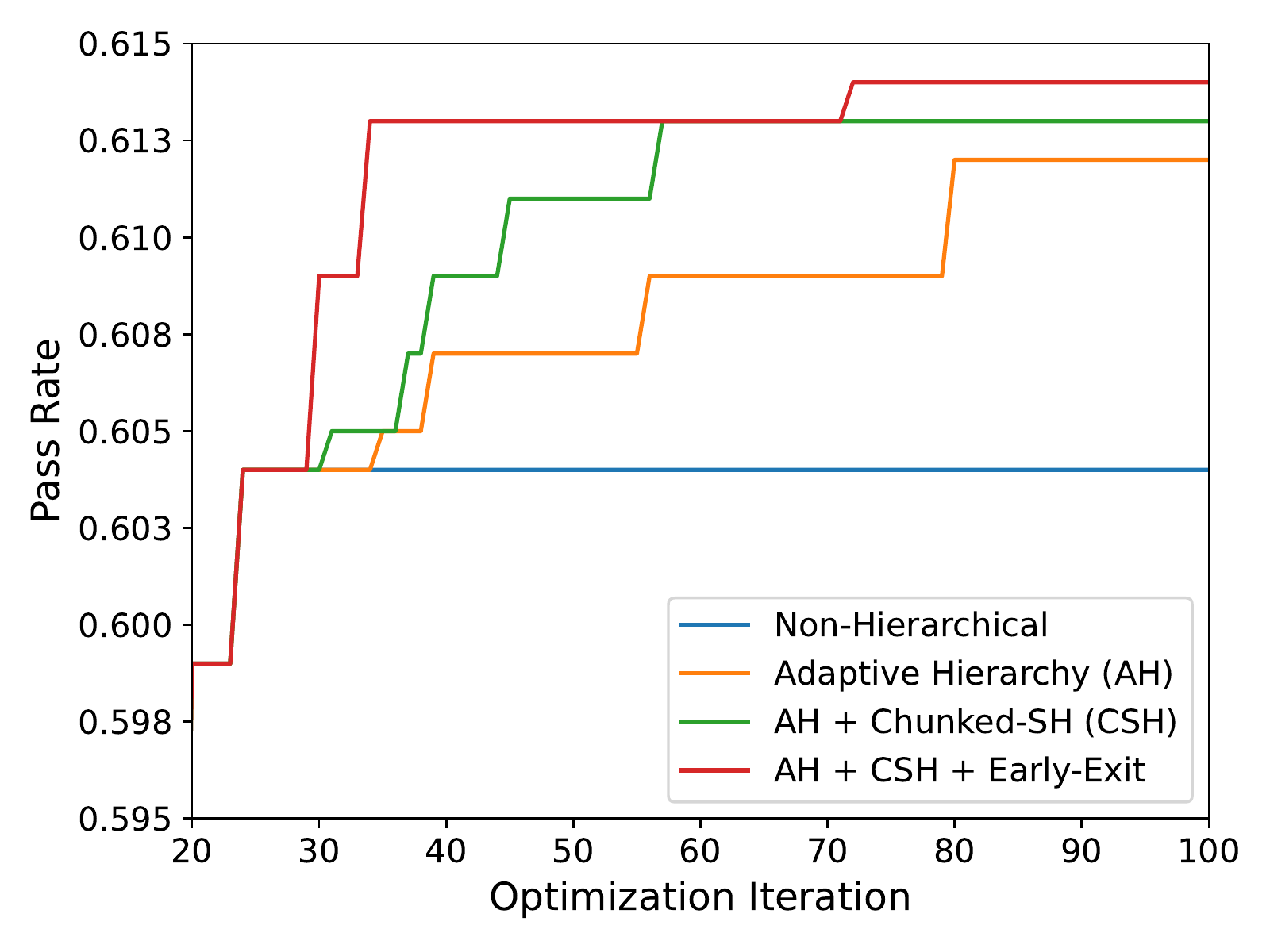}}
\vspace{-0.1in}
\mycap{Ablation Study Results.}
{
Performed on the Text-to-SQL workflow. The original workflow's pass rate is 0.49.
}
\Description{A line graph with multiple lines, each representing an added optimization.}
\label{fig-ablation-study}
\end{center}
\end{figure}
}

%% file: related.tex
\section{Related Works}
\label{sec:related}
This section discusses works related to us.

\input{fig-opt-cost}

\paragraph{Gen-AI workflow developing frameworks.}
Recent years have seen a surge of programming frameworks that facilitate the development of gen-AI workflows, such as LangChain~\cite{langchain-repo}, LlamaIndex~\cite{llamaindex}, OpenAI Swarm~\cite{swarm}, CrewAI~\cite{crewai}, and Dify~\cite{Dify}, Vellum~\cite{vellum}, and Coze~\cite{coze}.
These programming frameworks allow programmers to more easily develop and test their workflows, but does not offer workflow autotuning capabilities.
\sysname's design is compatible with these frameworks, as it can be used as a tool after developers write their programs with these frameworks. For example, \sysname\ currently supports out-of-the-box LangChain and DSPy programs.

\paragraph{Gen-AI workflow autotuning systems.}
While this paper provides the first comprehensive formalization and solution for gen-AI workflow autotuning, there are a few other works targeting the optimization of gen-AI workflows, primarily LLM-based workflows, as summarized in Table~\ref{tbl-comparison}. As seen, \sysname\ is the first autotuning system that incorporates workflow structure change, allows for multiple optimization objectives, and is fully extensible.

Existing gen-AI workflow optimizers can be categorized into two groups based on their optimization approaches.
The first group relies on an LLM to propose workflow changes and guide workflow autotuning.
For example, OPRO~\cite{opro} and Agent Symbolic Learning (Symbolic)~\cite{symbolic} use LLMs to directly refine prompts of language model calls in a workflow. TextGrad~\cite{textgrad} lets an LLM evaluate the result of a workflow with an LLM-generated ``loss'' and asks an LLM to improve prompts at different LM call sites based on the loss (``backpropagating'' the textual feedback). Trace~\cite{Trace} extends this concept of LLM-based backpropagation to let LLMs rewrite user-annotated code blocks. Different from these works, \sysname\ takes a data-driven approach; its workflow optimization is based on the sampled evaluation of workflow end results instead of asking the LLM for feedback. While an LLM can be useful in proposing improvements to the workflow, it is less stable as a feedback mechanism, as shown by our superior results than Trace. 

The second group searches over optimization options guided by workflow evaluation results.
DSPy~\cite{khattab2024dspy,dspy-2-2024,DSPy-repo} is a gen-AI workflow programming and optimization framework that applies various prompt tuning techniques like adding few-shot examples and CoT prompts for improving workflow generation quality. It supports several variations of OPRO as the search optimizer~\cite{dspy-2-2024}. Unlike \sysname, DSPy does not adapt their search according to total budgets and only focuses on prompt tuning for higher quality. GPTSwarm~\cite{gptswarm} optimizes DAG workflows by iteratively updating nodes and edges using the REINFORCE algorithm~\cite{reinforce}. \sysname\ supports generic graphs, including ones that contain cycles, and supports step changes. Furthermore, \sysname\ adapts to limited budgets, whereas GPTSwarm requires orders of magnitude more optimization iterations due to its use of reinforcement learning. 

\paragraph{Single model call optimizers.} There are several optimizers for a single call to gen-AI models. For example, RouteLLM~\cite{routellm} and TensorOpera-Router~\cite{tensor-opera-router} train a model to route LLM requests to a more cost-effective model. 
FrugalGPT~\cite{frugalgpt} sequentially retries a request with more expensive models until a particular score threshold is met. 
Differently, \sysname\ targets the optimization of an entire workflow, where optimizing steps in isolation does not efficiently or effectively work at the workflow level.

%% file: fig-opt-cost.tex
{
\begin{figure}[h]
\centerline{\includegraphics[width=\columnwidth]{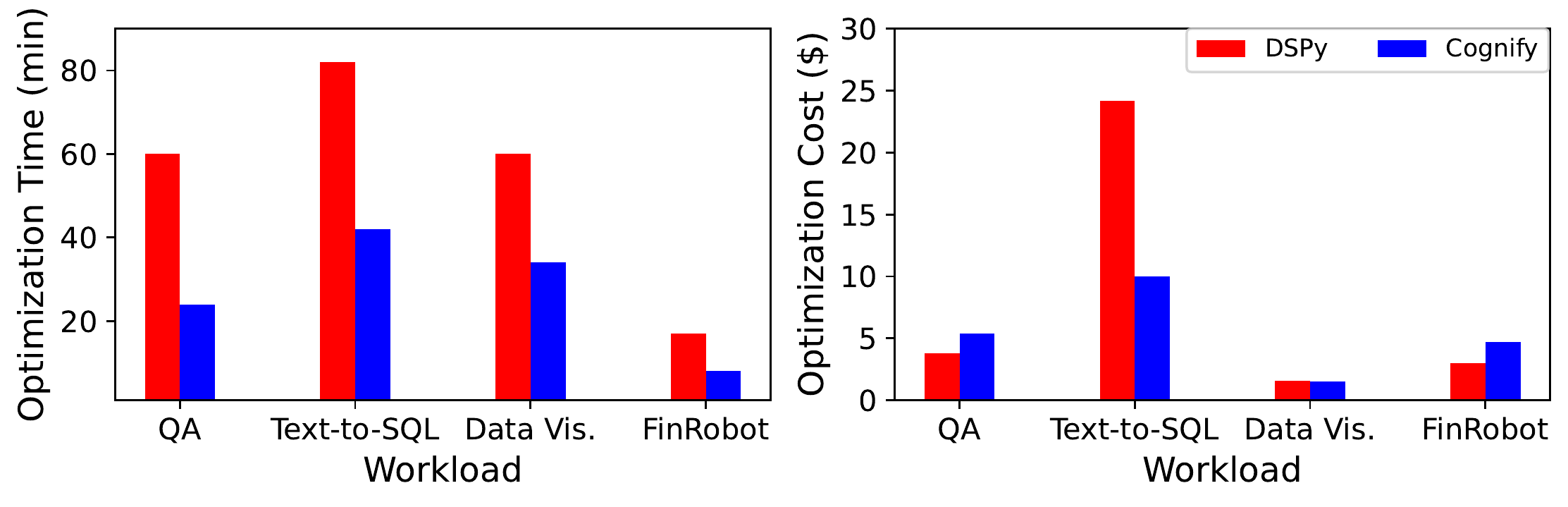}}
\vspace{-0.1in}
\mycap{Search Time (left) and Cost (right).} {Comparison between DSPy and Cognify (Trace omitted due to poor local convergence)}
\Description{Two bar graphs comparing a costly baseline to the optimized version.}
\label{fig-opt-time-cost}
\end{figure}
}

%% file: conclude.tex
\section{Conclusion}
\label{sec:conclude}

This paper formalizes the problem of gen-AI workflow autotuning and outlines the key differences between it and traditional ML autotuning/training. Based on our insights, we proposed an adaptive, hierarchical BO-based optimization algorithm, \search\ and demonstrated its robustness. We presented \sysname, a multi-objective gen-AI workflow autotuning platform we built on top of \search. Our evaluation results show that \sysname\ significantly pushes the Pareto frontier on generation quality, cost, and latency for workflows across a wide range of domains. We also demonstrated the robustness of the \search\ algorithm, allowing \sysname\ to serve as a generic, extensible, open-source gen-AI workflow autotuning platform future researchers and practitioners can leverage.

\begin{acks}
We would like to thank Jingbo Shang, Arya Gaikwad, and Yutong Huang from UCSD for their contribution and feedback to this paper.
\end{acks}

%% file: appendix.tex

\appendix
\section{Appendix}

\subsection{Bayesian Optimization}

\input{algo-bo}

Assuming our goal is to minimize the objective function $f(x)$. At each iteration $i$, Bayesian optimization (BO) constructs a surrogate model $S := p(y \mid x, D)$ to approximate the objective function $f$, where the set $D = \{(x_n,y_n)\}^{i-1}_{n=0}$ contains previously observed data points. BO also leverages an acquisition function to balance exploration and exploitation, based on predictions from $S$. The optimization process is be summarized in ~Alg \ref{alg:bo}. A widely used acquisition function is \textbf{Expected Improvement (EI)} \cite{Jones1998EfficientGO}:

\begin{align*}
    \text{EI}_{y^*}(x) := \int_{-\infty}^{y^*}(y^* - y)p(y\mid x, D)dy
\end{align*}

EI intuitively indicates the expectation that the objective function $f(x)$, approximated by $S$, will be below a certain threshold $y^*$.

\subsubsection{Optimizing EI with Tree-structured Parzen Estimator (TPE)}
\label{appdx:TPE}
In this work, we choose TPE as our surrogate model for its natural support of mixed discrete and continuous search space and efficiency in model building. TPE ~\cite{NIPS2011_TPE} employs kernel density estimators (KDEs) as the probabilistic model. Instead of modeling $p(y\mid x, D)$ directly, it models $p(x\mid y, D)$ by:

\begin{align*}
    p(x\mid y, D) := \begin{cases}
      l(x) := p(x\mid D^{(l)}) & y \leq y^{\pi}\\
      g(x) := p(x\mid D^{(g)}) & y > y^{\pi}
    \end{cases}       
\end{align*}

where the entire observation set $D$ is separated into two groups and $y^{\pi}$ is the top $\pi$ quantile evaluation results in $D$. The control parameter $\pi$ is calculated based on $|D|$ in each iteration ~\cite{Watanabe2023TreestructuredPE}. The next configuration $x^*$ to evaluate is proposed with:
\begin{align*}
    x^* := \argmax_{x \in \chi} \frac{l(x)}{g(x)}
\end{align*}

Intuitively this represents the configuration that maximizes the probability of residing in top-$\pi$ observations versus in worse group. ~\cite{bergstra2011tpe} has shown this is theoretically equivalent to minimizing EI. In \sysname, we also add observations that violate metric threshold to group $D^{(g)}$ so that the surrogate model can reason about the constraints.

\subsection{Budget Constraints}
Here we show that each layer $l$ will follow the assigned budget $B_l$ in ~\ref{alg:outer}. Starting from the outer-most layer. The number of proposed configurations at this layer is $C_1 = KW \leq B_1$. The total number of budget consumed at the next layer is:

\begin{align*}
    C_2 &= \sum_{k}^K \sum_{s}^S r_s\cdot|\Theta_s|\\
    &\leq\sum_{k}^K\sum_{s}^S R\eta^s \cdot W\eta^{-s} \\
    &\leq \frac{B_1}{W} \cdot \frac{B_{2}}{R} \cdot RW
\end{align*}
which aligns with the constraint of $B_1\cdot B_{2}$. The calculation can be easily extended to three-layer situation, where multiple instances of the search at the middle layer will be executed each with an uneven budget. The total number of evaluation required is:

\begin{align*}
    c_3[i] &\leq c_{2}[i]\cdot B_3\\
    C_3 &= \sum_i c_{3}[i] \\
    &\leq B_{3} \sum_i c_2[i] \\
    &\leq B_3 \prod_{j=1}^2B_j
\end{align*}
where $c_l[i]$ represents the budget assigned to $i^{th}$ optimize instance at search layer $l$. Sum of each $c_l[i]$ at layer $i$ stands for the total number of configurations proposed at that layer, which will be the product of all preceding budgets.

\subsection{Programming Model}
\label{sec:programming-model}

\sysname\ natively supports programs written in LangChain/LangGraph and DSPy. In addition, \sysname\ comes with its own programming model: \texttt{cognify.Model} and \texttt{cognify.StructuredModel}. These two constructs serve as drop-in replacements for all model calls in a user's program. There are four necessary components to construct a \texttt{cognify.Model}:
\begin{enumerate}
    \item \textbf{System prompt}: this defines the role of the model
    \item \textbf{Inputs}: placeholder values that are filled in at runtime based on the end-user request
    \item \textbf{Output format}: either a label assigned to the output value or a schema 
    \item \textbf{LM backend}: the model that will be called (in the absence of model selection)
\end{enumerate}

To invoke the optimizer, users simply need to register their workflow entry point, data loader, and evaluator functions with Cognify. The following snippet is an example of a workflow (defined in \texttt{workflow.py}).
\\
\\
\begin{lstlisting}[style=pythonstyle]
import cognify

math_solver = cognify.Model(
    "math_solver",
    system_prompt="You are a helpful assistant. Your task is to solve a math problem by identifying key variables and formulating important equations.",
    input_variables=[cognify.Input("problem")],
    output=cognify.OutputLabel("math_model"),
    lm_config=cognify.LMConfig(model="gpt-4o")
)

# Workflow
@cognify.register_workflow
def math_solver_workflow(problem):
    answer = math_solver(inputs={
        "problem": problem
    })
    return {"prediction": answer}
\end{lstlisting}

Then, in \texttt{config.py}, users can defined their data loader and evaluator.

\begin{lstlisting}[style=pythonstyle]
# Data loader
@cognify.register_data_loader
def load_data():
    with open("data.json", "r") as f:
        data = json.load(f)

   dataset = []
   for d in data:
      input_sample = {
         'problem': d["problem"],
      }
      label = {
         'label': d["solution"],
      }
      dataset.append(
        (input_sample, label)
      )

   # split the data into train, validation, and test
   return dataset[:20], dataset[20:30], dataset[30:]

# Evaluator
@cognify.register_evaluator
def score_solution(prediction, label):
    return prediction == label
\end{lstlisting}

Invoking the optimizer is straightforward.

\begin{lstlisting}[style=pythonstyle]
$ cognify optimize workflow.py
\end{lstlisting}

\subsection{Example Workflow}
\label{sec:apdx-example}

The following is the workflow used for the data visualization.
\input{fig-datavis}

An example optimization for this workflow is as follows:
\begin{itemize}
    \item Query expansion: Few-shot example (shown below)
    \item Initial code generation: Reasoning (with planning option, \ie, ``break down the task into simpler sub-tasks...") 
    \item Debugger: Reasoning (with chain-of-thought option, \ie, ``think step-by-step...")
\end{itemize}

The example used for query expansion was selected due to strong performance during the optimization process:\\

\begin{mdframed}[backgroundcolor=gray!20, linewidth=1pt]
\#\# Example 1 \#\#\\

\textbf{Query:} Create a combination chart from the ``data.csv" dataset, which contains sales data for various mobile phone brands across different quarters of the year. The columns in the CSV file are [``Quarter", ``Samsung", ``Nokia/Microsoft", ``Apple", ``LG", ``ZTE", ``Huawei"]. For each brand, create a box plot to represent the distribution of their sales data. On each box plot, display all the individual sales data points for that brand. Additionally, calculate the average sales for each brand and draw a line connecting these average values across the box plots. Use a consistent color scheme for the same quarter across different years, varying the shades to distinguish between years. Include a legend to aid in understanding the color coding.\\

\textbf{Output:} To create a combination chart from the ``data.csv" dataset that includes box plots for sales data of various mobile phone brands and a line connecting the average sales, follow these detailed instructions:

\#\#\# Step-by-Step Instructions\\
\#\#\#\# Step 1: Import Necessary Libraries\\
You will need to import the following libraries to handle data manipulation and visualization:
\begin{lstlisting}[style=pythonstyle]
import pandas as pd
import matplotlib.pyplot as plt
import seaborn as sns
\end{lstlisting}

\#\#\#\# Step 2: Load the Data\\
Read the CSV file into a pandas DataFrame. Ensure that the file ``data.csv" is in the same directory as your script or provide the full path.

\begin{lstlisting}[style=pythonstyle]
# Load the dataset
data = pd.read_csv('data.csv')
\end{lstlisting}

\#\#\#\# Step 3: Data Preparation\\
Check the structure of the DataFrame to ensure it has been loaded correctly. We will also need to calculate the average sales for each brand.

\begin{lstlisting}[style=pythonstyle]
# Display the first few rows of the dataset
print(data.head())

# Calculate the average sales for each brand
average_sales = data.mean()
\end{lstlisting}

\#\#\#\# Step 4: Create Box Plots\\
We will create box plots for each brand using seaborn. We will also overlay the individual sales data points on the box plots.

\begin{lstlisting}[style=pythonstyle]
# Set the style of seaborn
sns.set(style="whitegrid")

# Create a figure and axis
plt.figure(figsize=(12, 6))

# Melt the DataFrame to long format for seaborn
data_melted = data.melt(id_vars='Quarter', var_name='Brand', value_name='Sales')

# Create the box plot
box_plot = sns.boxplot(x='Brand', y='Sales', hue='Quarter', data=data_melted, palette='Set2', dodge=True)

# Overlay individual data points
sns.stripplot(x='Brand', y='Sales', hue='Quarter', data=data_melted, palette='Set2', dodge=True, marker='o', alpha=0.5)
\end{lstlisting}

\#\#\#\# Step 5: Plot Average Sales Line\\
Now we will plot the average sales for each brand as a line connecting the average values.

\begin{lstlisting}[style=pythonstyle]
# Create a line plot for average sales
for i, brand in enumerate(average_sales.index):
  plt.plot([i], [average_sales[brand]], marker='o', color='black', markersize=8, label=f'Avg {brand}' if i == 0 else "")
\end{lstlisting}

\#\#\#\# Step 6: Customize the Plot\\
Add titles, labels, and a legend to make the plot more informative.

\begin{lstlisting}[style=pythonstyle]
# Add titles and labels
plt.title("Sales Distribution of Mobile Phone Brands by Quarter")
plt.xlabel("Brand")
plt.ylabel("Sales")
plt.xticks(rotation=45)

# Add a legend
plt.legend(title="Quarter", bbox_to_anchor=(1.05, 1), loc='upper left')

# Show the plot
plt.tight_layout()
plt.show()
\end{lstlisting}

\#\#\#\# Step 7: Save the Plot (Optional)\\
If you want to save the plot as an image file, you can do so by adding the following line before plt.show():

\begin{lstlisting}[style=pythonstyle]
plt.savefig("sales_distribution_chart.png", bbox_inches='tight')
\end{lstlisting}

\#\#\# Complete Code\\
Here is the complete code for your reference:

\begin{lstlisting}[style=pythonstyle]
import pandas as pd
import matplotlib.pyplot as plt
import seaborn as sns

# Load the dataset
data = pd.read_csv('data.csv')

# Display the first few rows of the dataset
print(data.head())

# Set the style of seaborn
sns.set(style="whitegrid")

# Create a figure and axis
plt.figure(figsize=(12, 6))

# Melt the DataFrame to long format for seaborn
data_melted = data.melt(id_vars='Quarter', var_name='Brand', value_name='Sales')

# Create the box plot
box_plot = sns.boxplot(x='Brand', y='Sales', hue='Quarter', data=data_melted, palette='Set2', dodge=True)

# Overlay individual data points
sns.stripplot(x='Brand', y='Sales', hue='Quarter', data=data_melted, palette='Set2', dodge=True, marker='o', alpha=0.5)

# Calculate the average sales for each brand
average_sales = data.mean()

# Create a line plot for average sales
for i, brand in enumerate(average_sales.index):
  plt.plot([i], [average_sales[brand]], marker='o', color='black', markersize=8, label=f'Avg {brand}' if i == 0 else "")

# Add titles and labels
plt.title("Sales Distribution of Mobile Phone Brands by Quarter")
plt.xlabel("Brand")
plt.ylabel("Sales")
plt.xticks(rotation=45)

# Add a legend
plt.legend(title="Quarter", bbox_to_anchor=(1.05, 1), loc='upper left')

# Show the plot
plt.tight_layout()
plt.show()

# Optional: Save the plot as an image file
plt.savefig("sales_distribution_chart.png", bbox_inches='tight')
\end{lstlisting}

\#\#\# Conclusion\\
This code will generate a combination chart with box plots for each mobile phone brand's sales data, individual sales points, and a line connecting the average sales for each brand. Adjust the color palette and other parameters as needed to fit your specific requirements.

\end{mdframed}

%% file: algo-bo.tex
\begin{algorithm}[H]
\caption{Bayesian Optimization}
\label{alg:bo}
\begin{algorithmic}[1]
\STATE \textbf{Input:}\\
\hspace{1em}Evaluator function $f$\\
\hspace{1em}Search Budget $B$\\
\hspace{1em}Acquisition function $a$\\
\hspace{1em}Surrogate model $S$
\STATE \textbf{Define:} History observations $D = \emptyset$
    \FOR{$t \in 0, \dots, B-1$}
        \STATE $S_t = \text{Fit}(D)$ \COMMENT{Get the surrogate model with available observations}
        \STATE $x^* = \argmin_{x\in \chi} a(S_t, D)$ \COMMENT{Sample next configuration to evaluate}
        \STATE $y^* = f(x^*)$
        \STATE $D = D \cup \{(x^*, y^*)\}$
    \ENDFOR
\STATE \textbf{Output:} $H$
\end{algorithmic}
\end{algorithm}

%% file: fig-datavis.tex
{
\begin{figure}[h]
\begin{center}
\centerline{\includegraphics[width=\columnwidth]{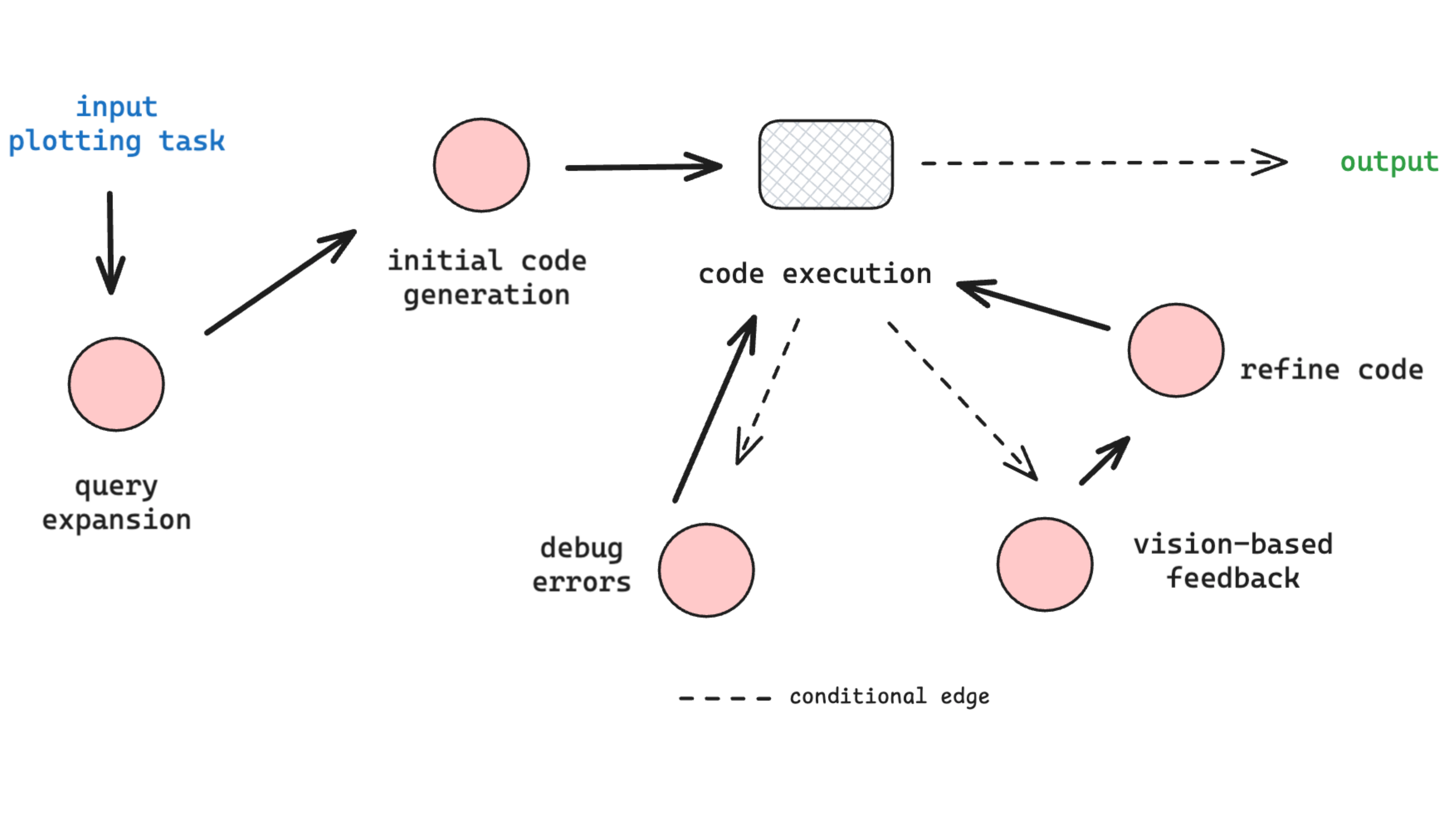}}
\mycap{Data Visualization Workflow}
{
Input is natural language description of a plotting task along with a CSV file containing relevant data. Expected output is a figure. Workflow contains 4 agents.
}
\end{center}
\Description{A task graph with 5 nodes and 2 conditional edges.}
\label{fig-datavis-workflow}
\end{figure}
}